%% file: main-paperID.tex
\documentclass[runningheads]{llncs}

\usepackage{graphicx}
\usepackage{comment}
\usepackage{amsmath,amssymb}
\usepackage[table]{xcolor}
\usepackage{capt-of}  
\usepackage{cuted}
\usepackage{color}
\usepackage{url}
\usepackage{hyperref}
\usepackage{xcolor} 
\usepackage[normalem]{ulem}

\newif\ifreview
\reviewtrue

\begin{document}
\input{text/our_macros}

\def\SubNumber{88}

\def\GCPRTrack{Fast Review Track}

\title{Neural Restoration of Greening Defects in Historical Autochrome Photographs Based on Purely Synthetic Data}

\ifreview
	 \author{%
    Saptarshi Neil Sinha\inst{1} \and
    Paul Julius K\"uhn\inst{1} \and
    Johannes Koppe\inst{2} \and
    Arjan Kuijper\inst{2} \and
    Michael Weinmann\inst{3}
  }
  \institute{%
    Fraunhofer IGD, Germany \and
    TU Darmstadt, Germany \and
    Delft Univeristy of Technology
  }
\else

	\author{First Author\inst{1}\orcidID{0000-1111-2222-3333} \and
	Second Author\inst{2,3}\orcidID{1111-2222-3333-4444} \and
	Third Author\inst{3}\orcidID{2222--3333-4444-5555}}
	
	\authorrunning{F. Author et al.}
	
	\institute{Princeton University, Princeton NJ 08544, USA \and Springer Heidelberg, Tiergartenstr. 17, 69121 Heidelberg, Germany
	\email{lncs@springer.com}\\
	\url{http://www.springer.com/gp/computer-science/lncs} \and ABC Institute, Rupert-Karls-University Heidelberg, Heidelberg, Germany\\
	\email{\{abc,lncs\}@uni-heidelberg.de}}
\fi

\maketitle              

\input{text/abstract}
\input{text/introduction}
\input{text/relatedworks}
\input{text/methodology}
\input{text/dataset}
\input{text/evaluation}
\input{text/conclusion}

%
%
%
%
\bibliographystyle{splncs04}
\bibliography{egbib}

\end{document}

%% file: text/our_macros.tex
\def\ie{i.e.\ }
\def\eg{e.g.\ }
\def\etal{et~al.\ }
\def\wrt{w.r.t.\ }


\newcommand{\todo}[1]{\textcolor{red}{#1}}
\newcommand{\nrevision}[2]{{\color{red}\sout{#1}}{\color{orange}\uwave{#2}}}
\newcommand{\finalrevision}[2]{{}{\color{black} #2}}
\newcommand{\mwrevision}[2]{{}{\color{black} #2}}

%% file: text/abstract.tex
\begin{abstract}
The preservation of early visual arts, particularly color photographs, is challenged by deterioration caused by aging and improper storage, leading to issues like blurring, scratches, color bleeding, and fading defects.
\finalrevision{}{Despite great advances in image restoration and enhancement in recent years, such systematic defects often cannot be restored by current state-of-the-art software features as available e.g. in Adobe Photoshop, but would require the incorporation of defect-aware priors into the underlying machine learning techniques. However, there are no publicly available datasets of autochromes with defect annotations.}
In this paper, we \finalrevision{}{address these limitations and} present the first approach \finalrevision{for}{that allows} the automatic removal of greening color defects in digitized autochrome photographs. 
\finalrevision{}{For this purpose, we introduce an approach for accurately simulating respective defects and use the respectively obtained synthesized data with its ground truth defect annotations to train a generative AI model with a carefully designed loss function that accounts for color imbalances between defected and non-defected areas}.
\finalrevision{Our main contributions include a method based on synthetic dataset generation and the use of generative AI with a carefully designed loss function for the restoration of visual arts. To address the lack of suitable training datasets for analyzing greening defects in damaged autochromes, we introduce a novel approach for accurately simulating such defects in synthetic data. We also propose a modified weighted loss function for the ChaIR method to account for color imbalances between defected and non-defected areas.}{}
\finalrevision{}{As demonstrated in our evaluation, our approach allows for the efficient and effective restoration of the considered defects, thereby overcoming limitations of alternative techniques that struggle with accurately reproducing original colors and may require significant manual effort.}
\finalrevision{While existing methods struggle with accurately reproducing original colors and may require significant manual effort, our method allows for efficient restoration with reduced time requirements.}{}
%
\keywords{Image restoration \and Synthetic data \and Deep learning \and Defect detection \and Visual arts \and Cultural heritage}
\end{abstract}
%
%

%% file: text/introduction.tex
\section{Introduction}
\label{sec:intro}

\begin{figure}[htb!]
\centering
  \includegraphics[width=0.75\linewidth]{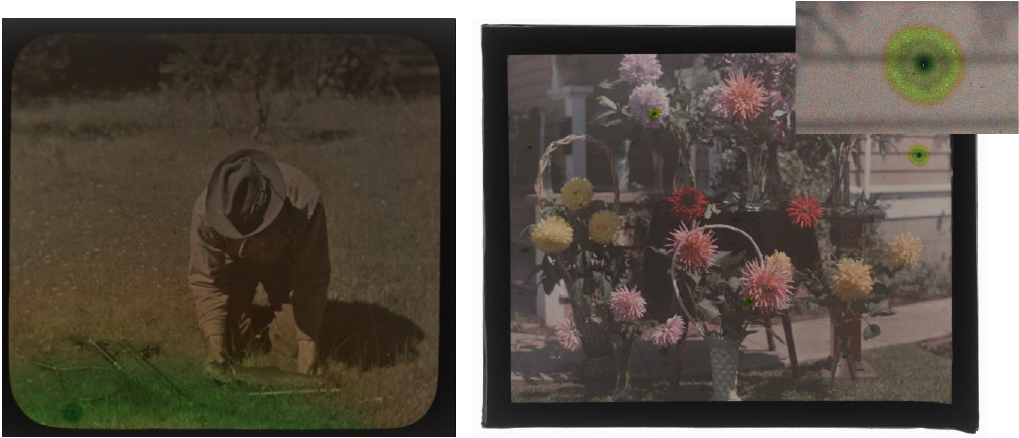}
  \caption{Greening defects in autochromes: (left) Large-region defects and (right) small green spots (highlighted with zoom-in) due to bleeding of dyes in the surrounding regions~\cite{lavedrine2013lumiere}.}
  \label{fig:greening_defects}
\end{figure}

Autochromes, invented in 1903 and introduced to the market in 1907 respectively by Auguste and Louis Lumière, represent the first widely and commercially adopted method for color photography~\cite{lavedrine2013lumiere}.
Utilizing glass plates coated with colored potato starch grains as color filters over a black-and-white emulsion, autochromes allowed for the creation of vivid and painterly color negatives through light projection, capturing the imagination of early 20th-century photographers and audiences.
\begin{figure}[htb!]
\centering
\includegraphics[width=\textwidth]{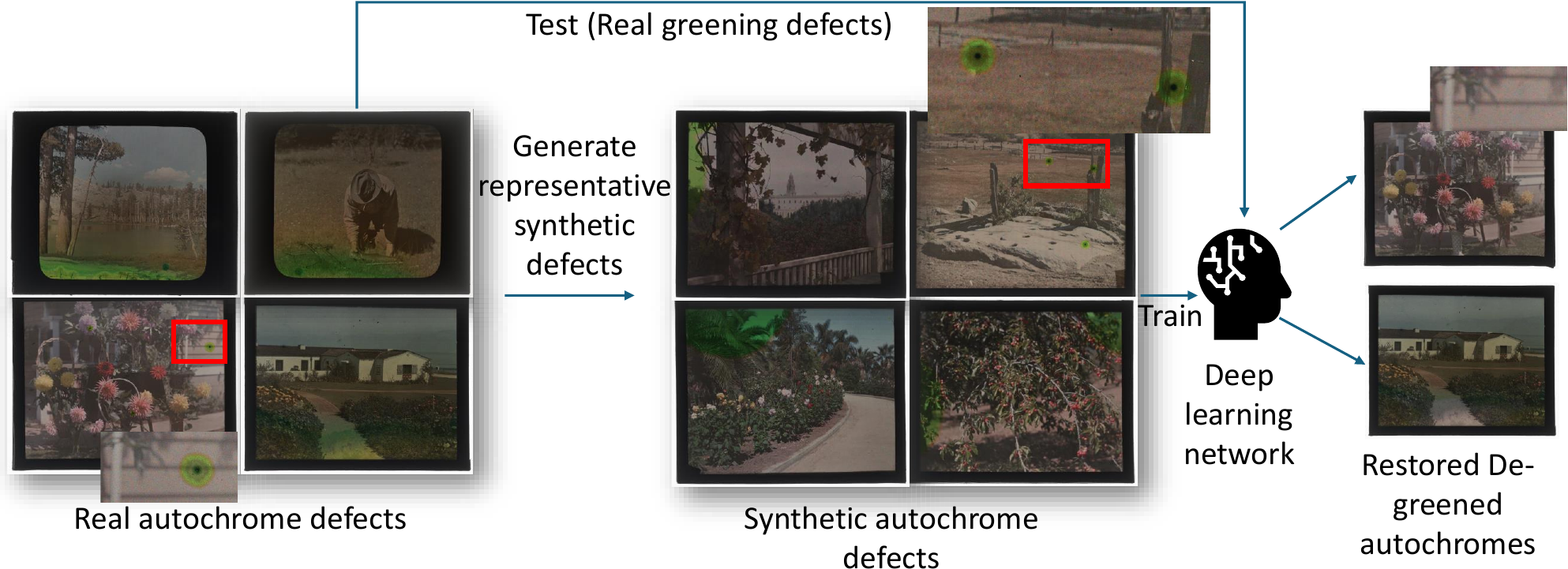} 
\captionof{figure}{\finalrevision{}{Pipeline for synthetic data generation and network training to restore greening defects in digitized autochromes.}}
\label{fig:pipeline}
\end{figure}
This principle dominated color photography for nearly three decades until it was replaced by Kodachrome films in the 1930s\finalrevision{which extended}{. The latter extended} autochromes by using multiple layers of light-sensitive emulsions, each sensitive to a specific color (red, green, or blue), and adding the dyes to the emulsion layers to reproduce the colors more accurately and offering finer grain and better resolution, similar to the principle of the Bayer filter in modern digital sensors.
In contrast to the archival stability of Kodachrome slides and films which can retain their colors for decades without significant fading \finalrevision{}{effects}, autochromes are fragile and sensitive to physical damage and environmental conditions.
Aging processes and inadequate storage may, hence, lead to deterioration in terms of blur, scratches, color bleeding and color fading. In particular, the dyes and emulsion layers are prone to fading or discoloration. 
Furthermore, the involvement of sensitive glass plate poses challenges for damage-free conservation, leading to various types of defects such as trapped dust, air bubbles, and moisture-related color issues.
A common defect, known as greening, occurs when the green potato starch grains bleed into adjacent areas due to their high sensitivity to water, resulting in unwanted green spots in the final image (see Fig. ~\ref{fig:greening_defects}).
Such artifacts distort the original appearance of these historical images, complicating their interpretation and diminishing their aesthetic and documentary value.
%
%

To address the aforementioned challenges, previous investigations on autochromes focused on the analysis of defects from delamination and faded dyes ~\cite{lavedrine2013lumiere} as well as the restoration of delamination ~\cite{Muller2006} and the restoration regarding faded dyes~\cite{Barker2023}.
Furthermore, the pre-trained real-ESRGAN model has been introduced to enhance old photographs, including autochromes, however, without specifically addressing greening defects~\cite{Saunders2021}. AI tools like Midjourney~\cite{Kovalev2024} and \finalrevision{}{the Generative Fill AI based inpainting tool in Adobe Photoshop~\cite{AdobePhotoshopGenerativeFill}}, aim to enhance autochromes, however, they also do not specifically address greening defects or utilize image pairs trained for this type of restoration. Hence, whereas modern deep learning methods, particularly image segmentation networks for defect detection and image restoration networks, offer promising solutions, the scarcity of reference data for historical artworks as well as the aforementioned systematic defects pose challenge for the training of models for restoring defects in autochromes like greening. \finalrevision{}{State-of-the-art image restoration methods, such as deblurring, denoising, super-resolution, shadow removal, desnowing, deraining, and dehazing~\cite{survey_IR}, are unsuitable for addressing greening defects in autochromes because they are not specifically trained or customized for this purpose. Furthermore, traditional restoration techniques like deblurring and denoising tend to remove the grainy texture that is essential for the aesthetics and authenticity of autochromes, which we aim to preserve. Additionally, methods such as super-resolution, shadow removal, desnowing, and deraining are primarily focused on restoring other image features and do not improve the treatment of greening defects. To \finalrevision{}{the best of} our knowledge, there is no existing work that effectively addresses the removal of greening defects in historical autochrome photographs using signal processing or deep learning techniques.}
%
%
\finalrevision{}{In this paper, address this research gap by presenting -- to the best of our knowledge -- the first successful learning-based approach for the automatic removal of green color defects in digitized autochrome photographs.}
For this purpose, we present the following main contributions:
\begin{itemize}
\setlength{\itemsep}{0.0em}
    \item An approach based on synthetic dataset generation and use of generative AI 
    with a carefully designed loss function for the restoration of visual arts (See Fig.~\ref{fig:pipeline}).
    \item To address the lack of suitable training datasets for analyzing greening defects in damaged autochromes, we present a novel approach for accurately simulating such defects in synthetic data.
    \item A modified weighted loss function for the ChaIR~\cite{chair} method to account for color imbalances between defected and non-defected areas. 
\end{itemize}
We will provide the code and dataset upon acceptance to facilitate further research, enhance transparency and reproducibility, and address ethical concerns~\cite{synthetic_ethical} that are important for synthetic data generation.

%% file: text/relatedworks.tex
\section{Related work}
\label{sec:related_work}
\noindent\textbf{Generative AI based image color restoration:}
Digital image processing can assist in artwork restoration, serving as a guide for traditional methods or enabling fully digital restorations~\cite{ImageProcessing_1,Kumar2024}. Recent generative methods, such as inpainting techniques, allow for masking and replacing areas or filling in missing parts~\cite{cm-gan,lama}. Other approaches modify the color of old photographs, including GAN-based colorization methods like ChromaGAN~\cite{vitoria2020chromagan}, CycleGAN~\cite{CycleGAN2017}, and Pix2Pix~\cite{pix2pix2017}. Additionally, models for white-balancing or color-balancing adjust color temperature globally~\cite{balancing_1,balancing_2,balancing_3}. Image decomposition methods separate layers, enabling applications like image deraining or dehazing~\cite{chair,decomposition_1,decomposition_2}, which help restore structures obscured by semi-transparent objects. The Channel Interaction Restoration (ChaIR) model~\cite{chair} achieves state-of-the-art performance on 13 benchmark datasets for dehazing, deblurring, and deraining by introducing a dual-domain channel attention mechanism that enhances interactions through lightweight convolutions in the spatial domain and integrates information from various frequency components. \finalrevision{This paper explores}{In this paper, we explore} the training, application, evaluation, and discussion of generative AI methods like ChaIR~\cite{chair}, CycleGAN~\cite{CycleGAN2017}, and Pix2Pix~\cite{pix2pix2017} for restoring greening defects in autochrome images, excluding diffusion-based approaches that tend to overwrite image structures due to their noise-based generation~\cite{limitation_diffusion}.
\finalrevision{We modify the loss function proposed in the ChaIR~\cite{chair} model so that it enhances the accuracy of color correction by specifically targeting defect areas, allowing for a more significant adjustment of color representations within these regions compared to the original model, which applied uniform corrections across the entire image.
}{In particular, we investigate whether modifying the loss function proposed in the ChaIR model~\cite{chair} allows enhancing the accuracy of color correction by specifically targeting defect areas, and demonstrate that this significantly outperforms the original ChaIR model due to a more significant adjustment of color representations within these regions compared to the original ChaIR model, which applied uniform corrections across the entire image.}
%

%
\noindent\textbf{Synthetic defect generation for visual arts:}
While multiple labeled datasets, such as the ART500K dataset~\cite{VA_dataset_1,VA_dataset_2}, are available for developing digital artwork restoration methods and encompass a variety of artworks from different painters and epochs, \finalrevision{challenges persist, including}{the still remaining challenges include} the scarcity of paired images of damaged and non-damaged artworks, the need to preserve colors and artistic styles, restoring large degraded areas with limited local information, generating masks for restoration areas, and identifying appropriate evaluation methods~\cite{Kumar2024}. A key factor in developing an effective AI model is the availability of a challenging and representative dataset. However, for artwork datasets in general and for our use-case of \finalrevision{}{addressing the removal of systematic} greening defects
occurring in autochromes
, there is insufficient annotated data for specialized image-to-image translation tasks. One solution is to create synthetic datasets~\cite{synthetic_1,synthetic_2,sinha_synthetic_data}, which have gained importance in computer vision in recent years, allowing for an unlimited amount of training data and facilitating faster automatic data labeling. 
\finalrevision{}{Exemplary works demonstrating the potential of synthesized training data for diverse applications include texture classification~\cite{targhi:2008}, material recognition~\cite{weinmann:2014}, pedestrian detection~\cite{vazquez:2012}, pedestrian classification~\cite{enzweiler:2008}, human pose recognition~\cite{shotton:2011}, the inference of intrinsic scene properties~\cite{Barron:2016} or semantic segmentation~\cite{richter:2016}.}
However, the generation and use of synthetic data can raise ethical and social concerns, as well as security and compliance issues, which is why it is essential to document these processes transparently~\cite{synthetic_1}. 
%
%
We address these concerns by incorporating expert knowledge regarding defects provided by an expert for autochrome defects to adequately mimic the defects in the scope of synthesized data, thereby overcoming the lacking availability of  ground truth (GT) data or image pairs including greening defects in old digitized autochrome images. In turn, the respective synthetic data allows the training of restoration models that can handle greening defects. 
%

%% file: text/methodology.tex
\section{Methodology}
\label{sec:methodology}
In this section, we introduce our core contributions on the generation of accurate, synthetic data including greening defects as required for the training of powerful restoration networks. Furthermore, we present extensions to restoration networks \finalrevision{relevant for the restoration of}{that specifically address defect restoration for} autochromes.

\begin{figure}[htb!]
\centering
  \includegraphics[width=0.75\linewidth]{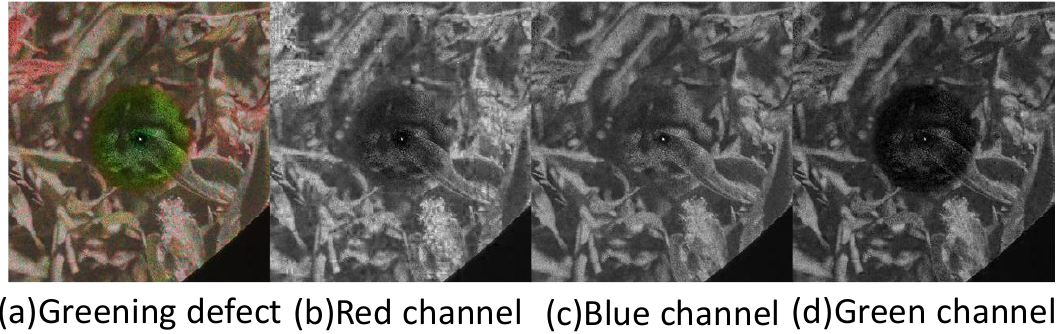}
  \caption{Channel composition of an example autochrome from the Harold Taylor collection~\cite{taylor1896}}
  \label{fig:channel_composition}
\end{figure}

\begin{figure}[htb!]
\centering
  \includegraphics[width=\linewidth]{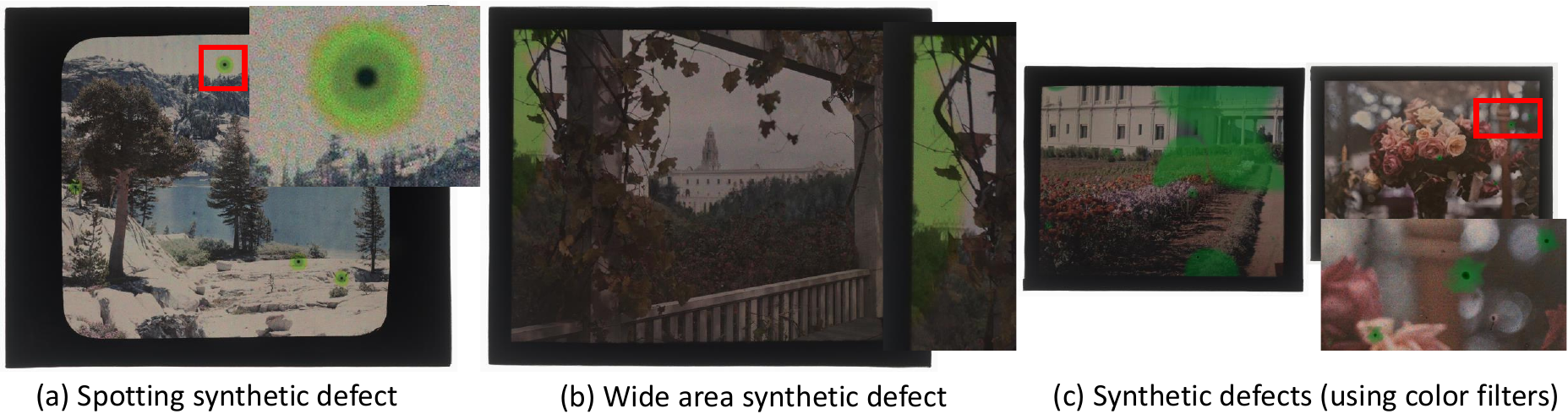}
  \caption{\finalrevision{}{(a) \& (b) --- Example synthetic greening defects using our synthetic defect geneation algorithm. (c) -- Synthetic defects generated using color filters.}}
  \label{fig:synthetic_defect}
\end{figure}

\noindent\textbf{Synthetic generation of greening defects:}
To address the key challenge of the absence of suitable datasets for damaged autochromes, we present an approach to create a synthetic dataset suitable for training and testing autochrome restoration methods.
%
To ensure the quality of the synthetic dataset, we closely examined real defects in autochromes
, revealing that they can be 
spot-shaped, wide-spread,
having both defect types, or completely damaged image.
\finalrevision{}{We selected 7 autochromes for closer evaluation, which revealed that spot defects have diameters ranging from 1\% to 5\% of the image width. Larger defects can occur individually, covering up to one-third of the image, although merging with other spot defects is rare.}
%
We also analyzed greening defects and their effects on individual color channels (see Fig.~\ref{fig:channel_composition}) 
based on a per-channel investigation
. Our findings revealed that the defects varied in color and transparency, were often bordered by orange tones, and appeared as dot-shaped rings with different shades of green and a dark core, originating from a single point and fading out in various directions due to liquid spreading.
%
Notably, the green color channel was the least affected, exhibiting increased intensity in damaged areas, whereas the red and blue channels showed significant decreases in color intensity (see Fig.~\ref{fig:channel_composition}).
%
%
In an initially generated dataset (\finalrevision{}{GreenFilterDefects}), we simulated defects based on color filters used in photography \finalrevision{}{(see Figure~\ref{fig:synthetic_defect} (c))}, specifically utilizing green filters (representing the foreground layer) for defect generation for an autochrome in the background layer. However, it turned out that this approach ultimately failed to produce realistic results, as the defects exhibited low color variance, making them appear unrealistic.
%
Hence, we created 
second dataset \finalrevision{}{(ChannelGreeningDefects)}
that directly represents observable changes in the color channels at the defective areas.
The process of defect generation is described below:
\begin{itemize}
\setlength{\itemsep}{0.0em}
    \item \finalrevision{}{Images may contain point defects (60\%), large defects (30\%), or both (10\%), with defect masks generated according to the characteristics of these known defects. Defect masks are created with 1 to 7 spot-shaped defects or 1 to 2 larger defects. The sizes and centers of the defects are again defined based on the known size ratios from real defects.}
    \item \finalrevision{}{The generation of area and spot-shaped defects relies on creating ellipses that are distorted by random radius adjustments at their boundaries. Spot defects have their centers located within the image, resulting in point-shaped origins. In contrast, large area defects involve generating significantly larger ellipses with origins outside the image, simulating the leakage of liquids that penetrate the autochrome and cause damage. The parametric equations for the ellipse are given by:
\begin{align*}
x(t) &= x_c + a \cdot \cos(t) \cdot \text{irregularity}, \\
y(t) &= y_c + b \cdot \sin(t) \cdot \text{irregularity},
\end{align*}
where:
\begin{itemize}
    \item $(x_c, y_c)$: co-ordinates of the center of the ellipse.
    \item $(a,b)$: Semi-major and semi-minor axis length of the ellipse.
    \item $t$: an angular parameter in $[0, 2\pi]$.
    \item \text{irregularity}: A factor affecting the shape of the ellipse.
\end{itemize}}
    \item \finalrevision{}{The center of defect masks may be point-shaped or linear. The intensity of the defect is interpolated based on the normalized distance \( d \) to the origin. The decrease in intensity is computed by $I = -d^2 + 1$.}
    \item\finalrevision{}{ For each ring, a change in percent of the individual color channels in the defect areas is defined. These
are randomly adjusted by a factor of 0.2 (chosen empirically to avoid drastic color channel changes and maintain realism of the defects) in both directions for each image. Table~\ref{tab:corruption_mapping}
shows the definition of the individual modifications per channel and ring in a dictionary}
    \item The combined defect pattern is smoothed with a Gaussian filter. The intensity change \(\Delta I_{c}\) for color channel \(c\) is calculated as \(\Delta I_{c} = (p_{c} * I_{c}) - I_{c}\), where \(p_{c}\) is the corruption percentage. This change is applied to the final image (\finalrevision{}{see Figure~\ref{fig:synthetic_defect} (a) \& (b)}).
\end{itemize}

\begin{table}[h]
    \centering
    \scalebox{0.75}{
    \begin{tabular}{|c|c|c|c|l|}
        \hline
        \textbf{Label} & \textbf{Blue} & \textbf{Green} & \textbf{Red} & \textbf{Effect in defect} \\ 
        \hline
        9   & 0.6   & 0.85  & 1.05 & Outer orange ring \\ \hline 
        1   & 0.5   & 1.2   & 0.8  & Light green ring     \\ \hline 
        2   & 0.4   & 0.8   & 0.6  & Middle               \\ \hline 
        3   & 0.4   & 0.8   & 0.6  & Middle               \\ \hline 
        4   & 0.2   & 0.6   & 0.1  & Dark green second    \\ \hline 
        99  & 0.2   & 0.2   & 0.1  & Dark mid             \\ \hline 
        20  & 0.4   & 0.95  & 0.6  & Surface              \\ 
        \hline
    \end{tabular}}
    \caption{\finalrevision{}{Dictionary for ring of corruption}}
    \label{tab:corruption_mapping}
\end{table}
%

\noindent\textbf{Training on restoration networks:} 
Being used for different use cases, we trained Image2Image models (Pix2Pix~\cite{pix2pix2017} and CycleGAN~\cite{CycleGAN2017}) on the synthetic data (\finalrevision{}{GreenFilterDefects}) with the original settings.
%
Furthermore, we trained a ChaIR model~\cite{chair} since it achieved state-of-the-art performance for deraining and dehazing tasks, on both of our datasets.
%
The original formulation relies on spatial and frequency loss functions to assess the model's performance in both domains, thereby supporting dual-channel attention mechanisms. Spatial and frequency loss function were defined as:
\begin{equation}
    l_s = \sum_{i=1}^{3} \frac{1}{S_i} \| \hat{Y}_i - Y_i \|_1
\end{equation}
\begin{equation}
    l_f = \sum_{i=1}^{3} \frac{1}{S_i} \| \mathcal{F}(\hat{Y}_i) - \mathcal{F}(Y_i) \|_1\
\end{equation}
where \(\hat{Y}_i\) stands for the predicted images and \(Y\) for the ground truth (GT) images, \(\mathcal{F}\) describes the Fast Fourier transform. The final loss is computed as \( l = l_s + 0.1 l_f\). \finalrevision{}{The weight 0.1 chosen in an empirical manner based on experiments that indicate it produces better results.} 
However, the main challenge for autochrome restoration is the correct representation of the original colors of the defective area and to place a stronger focus on the proper color representation of the defective area, we therefore replace the spatial loss \(l_{s}\) by:
\begin{equation}
l_s = \frac{1}{N} \sum_{x,y} W(x, y) \cdot | I_{\text{pred}}(x, y) - I_{\text{GT}}(x, y) |
\end{equation}
where,
\finalrevision{}{
\begin{equation}
W(x, y) = 
\begin{cases} 
1.0 & \text{if } |I_{\text{in}}(x, y) - I_{\text{GT}}(x, y)| > \text{t}, \text{ with } \text{t} = 0.1 \\
w & \text{otherwise, with } w \in \{0.1, 0.5\}
\end{cases}
\end{equation}}
The weight matrix W matches the size of the processed images and assigns a weight to incorrect color representations in defect areas that is two to ten times higher than in non-defective areas, while the remaining components of the loss function remain unchanged, with defect areas determined through a before-and-after comparison of input images and ground truth (GT) during training. \finalrevision{}{The value of the threshold parameter $t$ was determined empirically which best fits our dataset.}
%

%

%% file: text/dataset.tex
\section{Dataset and implementation}
\label{sec:datset}
\noindent 
We utilized the Harold Taylor collection~\cite{taylor1896}, comprising 420 autochrome images available under a public domain license for free use. We labeled the defects with assistance from an expert, identifying 306 images without visual damage and 95 with greening defects: 15 classified as having strong greening defects affecting a substantial part of the image, while the remaining 80 showed milder defects. We used the 306 defect-free images to synthetically generate defects, creating defected-undamaged pairs. Training was conducted on a cluster equipped with NVIDIA A100 GPUs. Corresponding training \finalrevision{}{configurations} are presented in \finalrevision{}{Table~\ref{tab:quantitative_results_v1} and Table~\ref{tab:quantitative_results_v2}.} 

%% file: text/evaluation.tex
\section{Evaluation}
\label{sec:evaluation}
\begin{figure}[htb!]
\centering
  \includegraphics[width=0.7\linewidth]{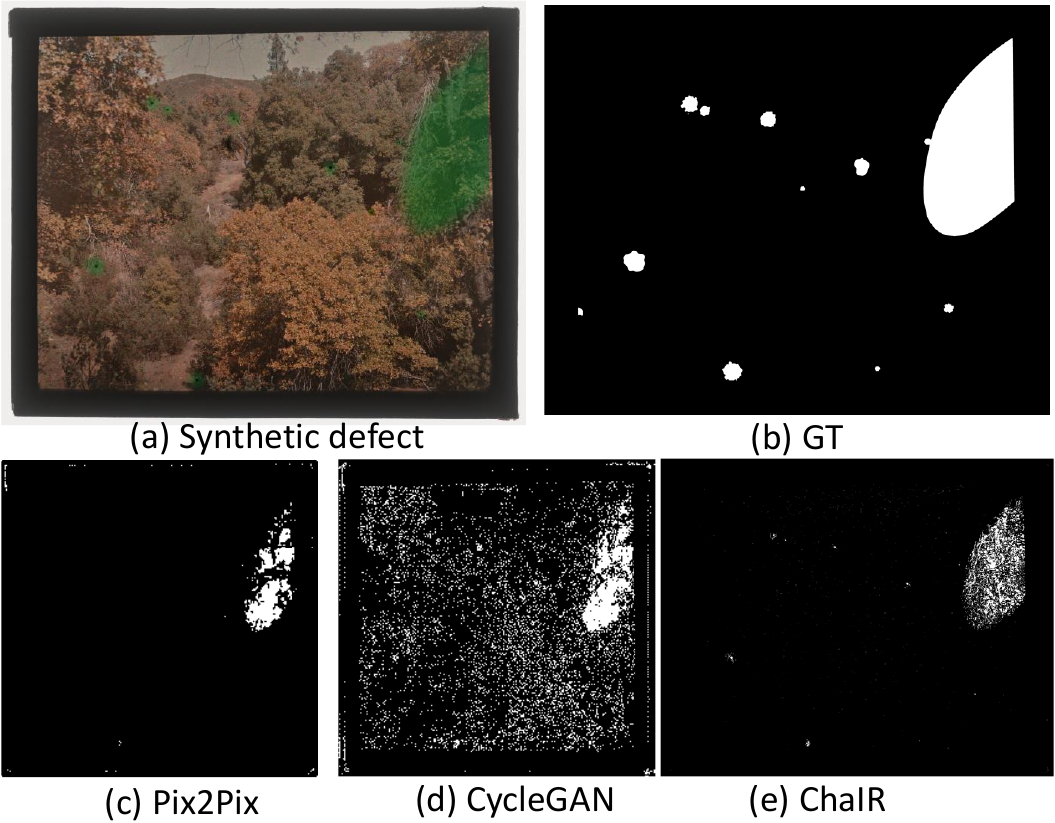}
  \caption{\finalrevision{}{Affected areas of the image after restored using models trained on GreenFilterDefects dataset}}
  \label{fig:affected_areas}
\end{figure}

To demonstrate the potential of our approach, we conducted both quantitative and qualitative analyses to evaluate the effectiveness of our method.
%
\begin{table}[htb!]
    \centering
    \caption{\finalrevision{}{Quantitative results (on GreenFilterDefects synthetic dataset (Figure~\ref{fig:synthetic_defect} (c)))}}
    \label{tab:quantitative_results_v1}
    \scalebox{0.85}{
    \begin{tabular}{|c|c|c|c|c|c|}
        \hline
        \textbf{Method} & \textbf{Dataset} & \textbf{Epochs} & \textbf{Loss Type} & \textbf{Mean PSNR} $\uparrow$ & \textbf{Mean MS-SSIM} $\uparrow$ \\ \hline
       pix2pix~\cite{pix2pix2017} & GreenFilterDefects & LR &  vanilla &  \cellcolor{orange!40}35.088 &  \cellcolor{yellow!40}0.983 \\ \hline
        cycleGAN~\cite{CycleGAN2017} & GreenFilterDefects & LR & vanilla & 29.311 & 0.959 \\ \hline
        PretrainedChaIR~\cite{chair} & RI ~\cite{DatasetPretrained} & 300 & - & 19.746 & 0.928 \\ \hline
        PretrainedChaIR~\cite{chair} & RO ~\cite{DatasetPretrained} & 30 & - & 22.216 & 0.947 \\ \hline
        ChaIR~\cite{chair} & GreenFilterDefects & 300 & S+F &  \cellcolor{yellow!40}34.333 & \cellcolor{orange!40}0.987 \\ \hline \hline
    \end{tabular}}
    \footnotesize{* S+F = spatial + frequency, RI = RESIDEIndoor, RO = RESIDEOutdoor, LR = 100 (initial) + 100 (decay), MS-SSIM=Multiscale SSIM, \\ \finalrevision{}{Loss function weights defected areas similarly as non-defected areas ($W(x,y)=1.0$)}}
\end{table}
\finalrevision{}{Initially, we performed our evaluation (see Table~\ref{tab:quantitative_results_v1}) on the GreenFilterDefects dataset (based on green color filters see Figure~\ref{fig:synthetic_defect} (c)) which showed the Image2Image models (cycleGAN~\cite{CycleGAN2017} and pix2pix~\cite{pix2pix2017}) affected regions outside the region of interest in contrast to channel based models like ChaIR~\cite{chair} as shown in Figure~\ref{fig:affected_areas}. Therefore, for the ChannelGreeningDefects dataset generated based on our algorithm as described in Section~\ref{sec:methodology}, we used the ChaIR model~\cite{chair}, by training from scratch as well as finetuning with the pretrained model (see Table~\ref{tab:quantitative_results_v2}).}
\begin{table}[htb!]
    \centering
    \caption{\finalrevision{}{Quantitative results (ChannelGreeningDefects synthetic dataset generated using our algorithm based on color channels (see Figure~\ref{fig:synthetic_defect} (a) \& (b)))}}
    \label{tab:quantitative_results_v2}
    \scalebox{0.72}{
    \begin{tabular}{|c|c|c|c|c|c|}
        \hline
        \textbf{Method} & \textbf{Dataset} & \textbf{Epochs} & \textbf{Loss Type} & \textbf{Mean PSNR} $\uparrow$& \textbf{Mean MS-SSIM} $\uparrow$\\ \hline
        ChaIR~\cite{chair} & ChannelGreeningDefects & 300 & S+F &34.795 & 0.985 \\ \hline
        ChaIRFinetunedV2 (Ours) & RO + ChannelGreeningDefects & 300 & S+F & \cellcolor{orange!40}39.492 & \cellcolor{orange!40}0.995 \\ \hline
    \end{tabular}}
    \footnotesize{* S+F = spatial + frequency, RO = RESIDEOutdoor,  MS-SSIM=Multiscale SSIM}
\end{table}
\begin{table}[htb!]
    \centering
    \small
    \caption{\finalrevision{}{Comparison with photo editing tools (Photoshop).}}
    \label{tab:photoshop}
    \scalebox{0.99}{
    \begin{tabular}{|c|c|c|}
        \hline
        \textbf{method} & \textbf{MS-SSIM} $\uparrow$ & \textbf{SSIM Cropout} $\uparrow$ \\ \hline
        GenFill\_Vanilla~\cite{AdobePhotoshopGenerativeFill} & 0.991 & 0.504 \\ \hline
        GenFill\_ManualCorrection & 0.996 & 0.946 \\ \hline
        ChaIR~\cite{chair} & \cellcolor{yellow!40}0.997 & \cellcolor{yellow!40}0.964 \\ \hline
        ChaIRFinetunedV2 (Ours) & \cellcolor{orange!40}0.998 & \cellcolor{orange!40}0.972 \\ \hline
    \end{tabular}}
       \footnotesize{\\MS-SSIM=Multiscale SSIM, SSIM = SSIM of the affected cropped regions}
\end{table}
\noindent\textbf{Quantitative results:} \finalrevision{}{The results in Table~\ref{tab:quantitative_results_v1} and Table~\ref{tab:quantitative_results_v2} present PSNR and SSIM scores between synthetic references and outputs, divided into two groups based on datasets GreenFilterDefects and ChannelGreeningDefects for comparability. Metrics are averaged over all test images. We see that the ChaIR~\cite{chair} performed best in the GreenFilterDefects dataset and 
ChaIRFinetuned~\cite{chair} performed best against the ChannelGreeningDefects dataset. Our loss function weights defected areas similarly as non-defected areas for these results (see Table~\ref{tab:quantitative_results_v1}).}
We also compared our results \finalrevision{}{with Generative Fill AI-based inpainting tools~\cite{AdobePhotoshopGenerativeFill} integrated into Photoshop (\emph{GenFill\_Vanilla}) and manual corrections by an expert designer using the Generative Fill method (\emph{GenFill\_ManualCorrection})}. As shown in Table~\ref{tab:photoshop}, the ChaIR model trained on synthetic data outperforms both the results \finalrevision{}{using this state-of-the-art photo editing tool.}
\begin{figure}[htb!]
\centering
  \includegraphics[width=\linewidth]{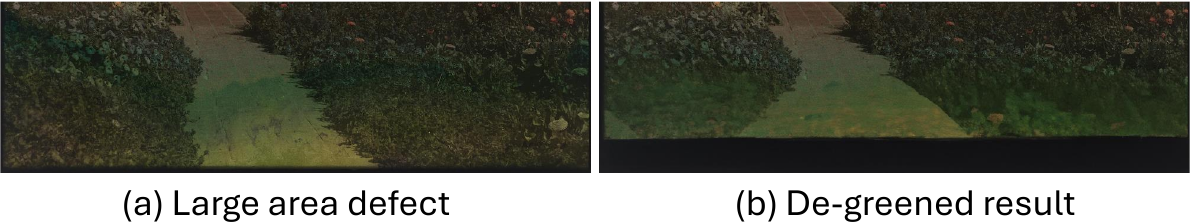}
  \caption{\finalrevision{Large area defects corrected using}{Limitation of state-of-the-art tools: Applying the} Generative Fill AI inpainting tool~\cite{AdobePhotoshopGenerativeFill} \finalrevision{}{does not successfully remove the greening artifacts.}}
  \label{fig:gen_fill_correction}
\end{figure}
\finalrevision{}{\noindent\textbf{Ablation study:}
The ablation study reveals that the implementation of different loss functions, particularly loss10 that is weighting the defected regions by 10, improves the defect correction compared to original loss in the ChaIR model~\cite{chair}.}
\begin{table}[htb!]
    \centering
    \caption{\finalrevision{}{Ablation study of the different loss functions}}
    \label{tab:ablation}
    \scalebox{0.82}{
    \begin{tabular}{|c|c|c|c|c|c|}
        \hline
        \textbf{Method} & \textbf{Dataset} & \textbf{Epochs} & \textbf{Loss Type} & \textbf{Mean PSNR} $\uparrow$ & \textbf{Mean MS-SSIM} $\uparrow$\\ \hline
        ChaIR~\cite{chair} & ChannelGreeningDefects & 300 & S+F & 34.795 & 0.985 \\ \hline
        ChaIRLoss2 (Ours) & ChannelGreeningDefects & 300 & ours (loss2) & \cellcolor{yellow!40}34.874 & \cellcolor{yellow!40}0.987 \\ \hline
        ChaIRLoss10 (Ours) & ChannelGreeningDefects & 300 & ours (loss10) & \cellcolor{orange!40}35.776 & \cellcolor{orange!40}0.989 \\ \hline
    \end{tabular}}
    \footnotesize{* S+F = spatial + frequency, ours (loss2 \finalrevision{}{($W=0.5$)}) and ours (loss10 \finalrevision{}{($W=0.1$)})) = defect areas penalized by factor of 2 and 10 respectively }
\end{table}
\noindent\textbf{Qualitative analysis:}
\begin{figure}[htb!]
\centering
  \includegraphics[width=0.7\linewidth]{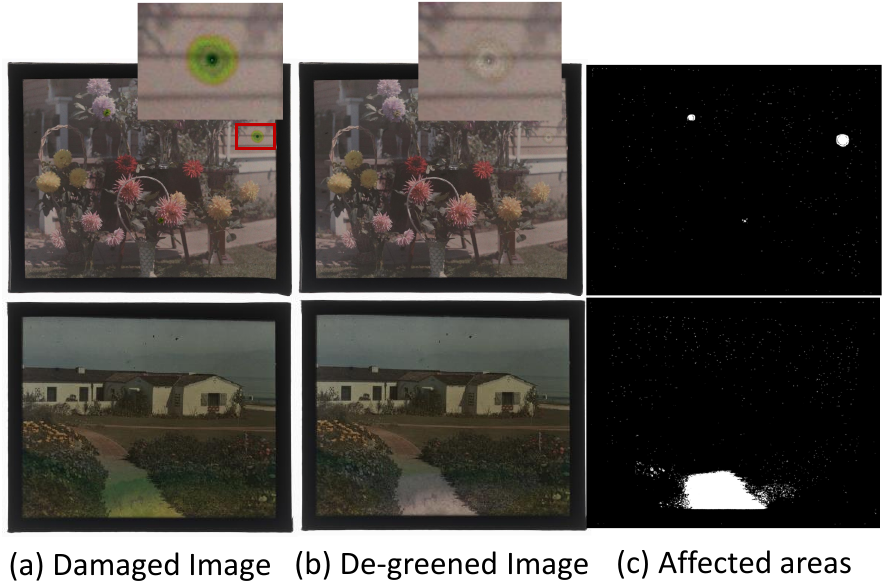}
  \caption{\finalrevision{}{Qualitative analysis on the real dataset (large and spotting defects), (b) --- shows the degreening (c) --- shows the affected regions of (a).}}
  \label{fig:qualitative_analysis}
\end{figure}
\finalrevision{}{ We performed a qualitative analysis of the spotting defects and large area real defects using the ChaIRFinetunedV2 model (see Figure~\ref{fig:qualitative_analysis}).} We selected this model due to its superior quantitative results. While greening defects are significantly reduced, small regions may still be inadvertently affected, particularly in images with larger defected areas. \finalrevision{}{The results indicate that the greening effect can be significantly reduced in larger areas (see Fig.~\ref{fig:qualitative_analysis}), whereas other state-of-the-art image restoration methods~\cite{survey_IR,instructIR}(see Fig.~\ref{fig:qualitative_analysis_large_area_IR}) and AI-powered photo-editing software~\cite{AdobePhotoshopGenerativeFill} are unable to achieve this(see Fig.~\ref{fig:gen_fill_correction}). Further evaluations on this matter are discussed below.}
\begin{figure}[htb!]
\centering
  \includegraphics[width=0.85\linewidth]{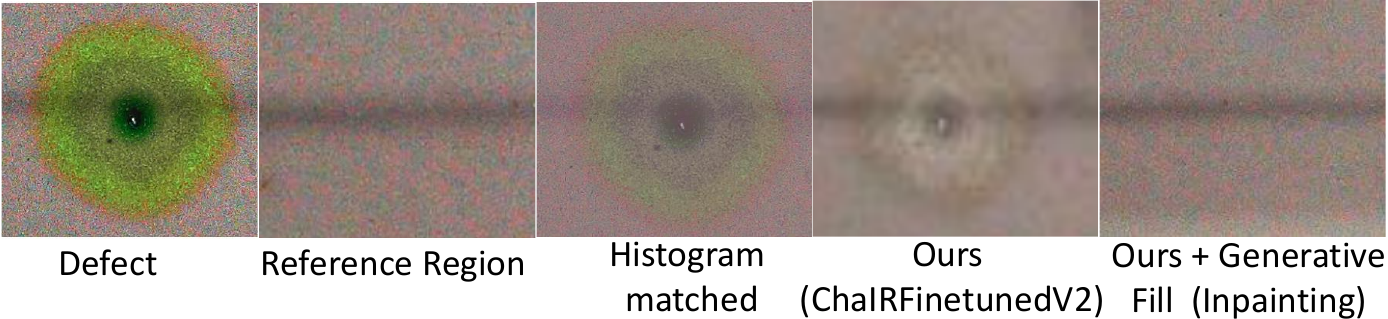}
  \caption{\finalrevision{}{Comparison of our de-greened result, histogram matching of the region of interest with the reference region and de-greened result followed by AI based inpainting}}
  \label{fig:histogram_matching}
\end{figure}
\finalrevision{}{\noindent\emph{\textbf{Comparison with state-of-the-art photo editing tool:}} The results of the 'Generative Fill' ~\cite{AdobePhotoshopGenerativeFill} AI-based inpainting tool (see Fig.~\ref{fig:gen_fill_correction}) indicate a noticeable change in the structure of the defective areas, suggesting that this method is unsuitable for restoration as it ignores and overwrites underlying structures unlike our method.}
\finalrevision{}{\noindent\emph{\textbf{Comparison with a signal processing approach:}} Fig.~\ref{fig:histogram_matching} demonstrates that our method outperforms classical techniques like histogram matching around defected regions, where the characteristic noise of the autochrome is modified (smoothed) and the greening effect is not completely removed. In comparison our method preserves the grainy structure and also reduces the greening effect considerably and final output is plausible which can be then processed adequately by a curator.}
\finalrevision{}{\noindent\emph{\textbf{Comparison with learning-based image restoration methods:}} Our method outperforms state-of-the-art learning-based image restoration (IR) methods (see Figure~\ref{fig:qualitative_analysis_large_area_IR} and \ref{fig:qualitative_analysis_spot_IR}) incorporated in InstructIR~\cite{instructIR} through prompting. We utilized curated prompts (from both GPT-4 and real users) as suggested in the InstructIR~\cite{instructIR} to trigger the corresponding IR methods like deraining, dehazing, etc..}
\begin{figure}[htb!]
\centering
  \includegraphics[width=0.9\linewidth]{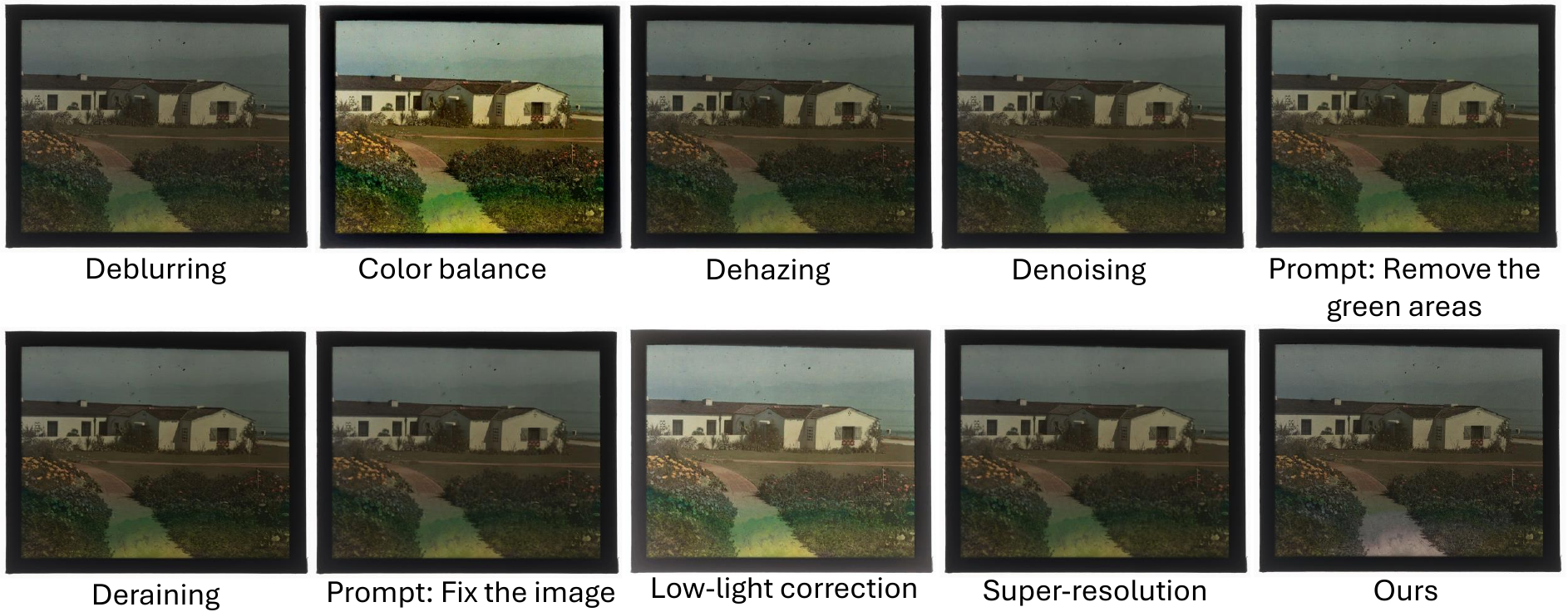}
  \caption{\finalrevision{}{Comparison of large area defects with learning-based image restoration approaches using InstructIR~\cite{instructIR}. The respective image see Figure.~\ref{fig:qualitative_analysis}}}
  \label{fig:qualitative_analysis_large_area_IR}
\end{figure}
\begin{figure}[htb!]
\centering
  \includegraphics[width=0.9\linewidth]{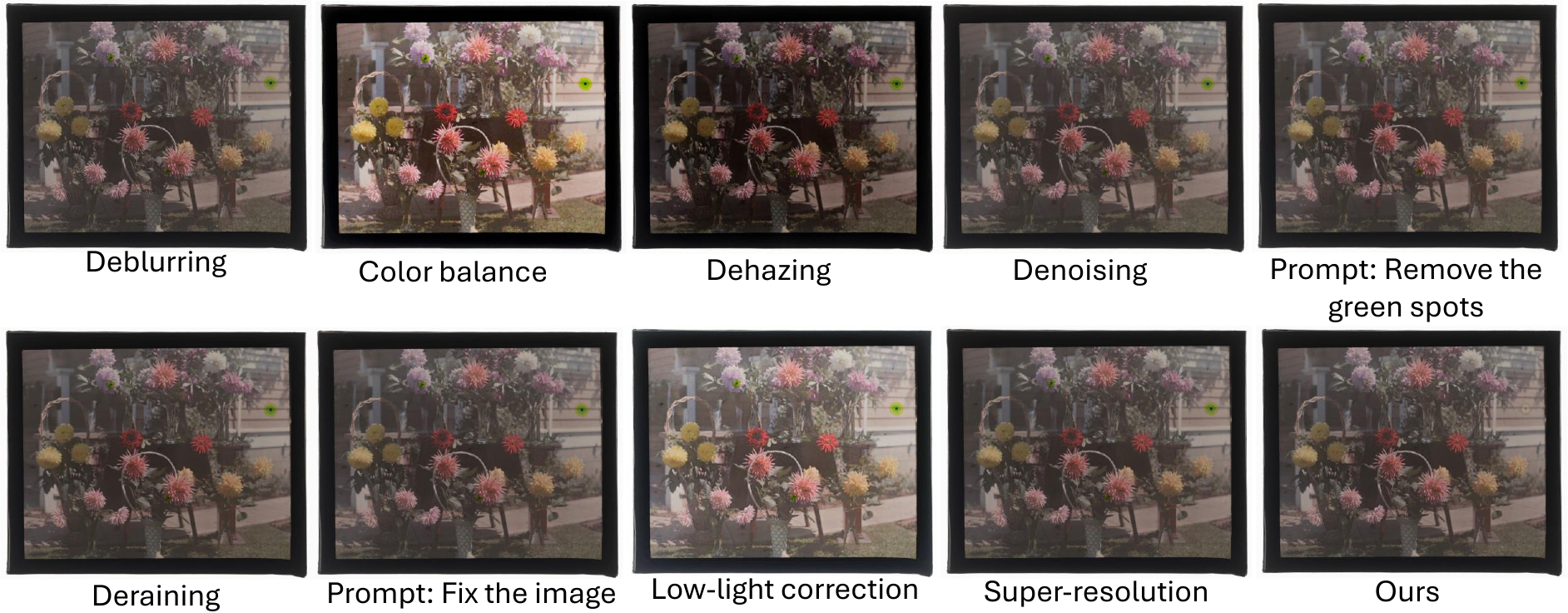}
  \caption{\finalrevision{}{Comparison of spotting defect with learning-based image restoration approaches using InstructIR~\cite{instructIR}. The respective defect image see Figure~\ref{fig:qualitative_analysis}.}}
  \label{fig:qualitative_analysis_spot_IR}
\end{figure}

%% file: text/conclusion.tex
\section{Conclusion}
\label{sec:conclusion}
Our method effectively identifies \finalrevision{}{greening} defects of all sizes and locations in old autochrome photographs and \finalrevision{}{corrects them} by adjusting the colors of detected areas, making them less recognizable. However, it sometimes struggles to accurately reproduce the original colors of autochrome images, often resulting in bluish tones and missing very small defects. \finalrevision{}{AI based inpainting methods enable targeted defect removal in stationary regions in small areas from our degreened results. Additionally, deep inpainting algorithms show promise in restoring greening defects on stationary areas using masks generated by our algorithms. State-of-the-art learning-based image restoration methods are unable to remove such defects.} 
Our method's output can enhance editing tools, allowing designers to achieve similar results more efficiently and tackle defects that are otherwise difficult to manage \finalrevision{}{especially in larger areas.} 
Future work should focus on expanding the synthetic dataset \finalrevision{}{to support digitized autochromes from other collections} and identifying suitable no-reference metrics \finalrevision{}{(metrics that evaluate image quality without needing a reference image for comparison, such as NIQE - Natural Image Quality Evaluator)} to evaluate restoration quality across a wider range of samples. Furthermore, the methodologies developed could be adapted to restore other autochrome defects, such as oranging, and \finalrevision{applied to various artworks and image types}{could be applied for the restoration of various artworks and image types}.
%

%